\documentclass[journal]{IEEEtran}

\newcommand{\ie}{\textit{i.e.}}
\newcommand{\eg}{\textit{e.g.}}

\usepackage{color}
\usepackage[switch,mathlines]{lineno}
\usepackage{amsmath,amsfonts}
\usepackage{algorithmic}
\usepackage{algorithm}
\usepackage{array}
\usepackage{bm}
\usepackage[caption=false,font=normalsize,labelfont=sf,textfont=sf]{subfig}
\usepackage{textcomp}
\usepackage{stfloats}
\usepackage{url}
\usepackage{verbatim}
\usepackage{graphicx}
\usepackage{cite}
\usepackage{multirow}
\usepackage{booktabs}
\usepackage{epstopdf}
\usepackage[capitalize]{cleveref}
\usepackage[flushleft]{threeparttable}
\usepackage{mathtools}
\DeclareMathOperator{\E}{\mathbb{E}}

\let\oldalign\align
\let\oldendalign\endalign

\renewenvironment{align}
  {\linenomathNonumbers\oldalign}
  {\oldendalign\endlinenomath}

\let\oldequation\equation
\let\oldendequation\endequation

\renewenvironment{equation}
  {\linenomathNonumbers\oldequation}
  {\oldendequation\endlinenomath}

\begin{document}

\title{E-MLB: Multilevel Benchmark for Event-Based Camera Denoising}

\author{Saizhe Ding\IEEEauthorrefmark{1}, Jinze Chen\IEEEauthorrefmark{1}, Yang Wang\IEEEauthorrefmark{2}, Yu Kang,~\IEEEmembership{Senior Member,~IEEE,} Weiguo Song, Jie Cheng and Yang Cao,~\IEEEmembership{Member,~IEEE,}
    \thanks{This work was supported by the National Natural Science Foundation of China (NSFC) under Grants 62206262, 62033012 and 52074252.}
    \thanks{\IEEEauthorrefmark{1} Saizhe Ding and Jinze Chen contributed equally to this paper.}
    \thanks{\IEEEauthorrefmark{2} Yang Wang is the corresponding author of this paper.}

    \thanks{Saizhe Ding and Weiguo Song are with State Key Laboratory of Fire Science, University of Science and Technology of China, Hefei, Anhui Province, China. (e-mail: dszh2020@mail.ustc.edu.cn, wgsong@ustc.edu.cn)}
    \thanks{Jinze Chen, Yang Wang, Yu Kang, Yang Cao are with School of Informatics, University of Science and Technology of China, Hefei, Anhui Province, China. (e-mail: chjz@mail.ustc.edu.cn, ywang120@ustc.edu.cn, kangduyu@ustc.edu.cn, forrest@ustc.edu.cn)}
    \thanks{Jie Cheng is with Huawei Technologies Co., Ltd. (e-mail: chengjie8@huawei.com)}}



\maketitle

\begin{abstract}
    Event cameras, such as dynamic vision sensors (DVS), are biologically inspired vision sensors that have advanced over conventional cameras in high dynamic range, low latency and low power consumption, showing great application potential in many fields. Event cameras are more sensitive to junction leakage current and photocurrent as they output differential signals, losing the smoothing function of the integral imaging process in the RGB camera. The logarithmic conversion further amplifies noise, especially in low-contrast conditions. Recently, researchers proposed a series of datasets and evaluation metrics but limitations remain: 1) the existing datasets are small in scale and insufficient in noise diversity, which cannot reflect the authentic working environments of event cameras; and 2) the existing denoising evaluation metrics are mostly referenced evaluation metrics, relying on APS information or manual annotation. To address the above issues, we construct a large-scale event denoising dataset (multilevel benchmark for event denoising, E-MLB) for the first time, which consists of 100 scenes, each with four noise levels, that is 12 times larger than the largest existing denoising dataset. We also propose the first nonreference event denoising metric, the event structural ratio (ESR), which measures the structural intensity of given events. ESR is inspired by the contrast metric, but is independent of the number of events and projection direction. Based on the proposed benchmark and ESR, we evaluate the most representative denoising algorithms, including classic and SOTA, and provide denoising baselines under various scenes and noise levels. The corresponding results and codes are available at {\url{https://github.com/KugaMaxx/cuke-emlb}}.
\end{abstract}

\begin{figure*}
    \includegraphics[width=\textwidth]{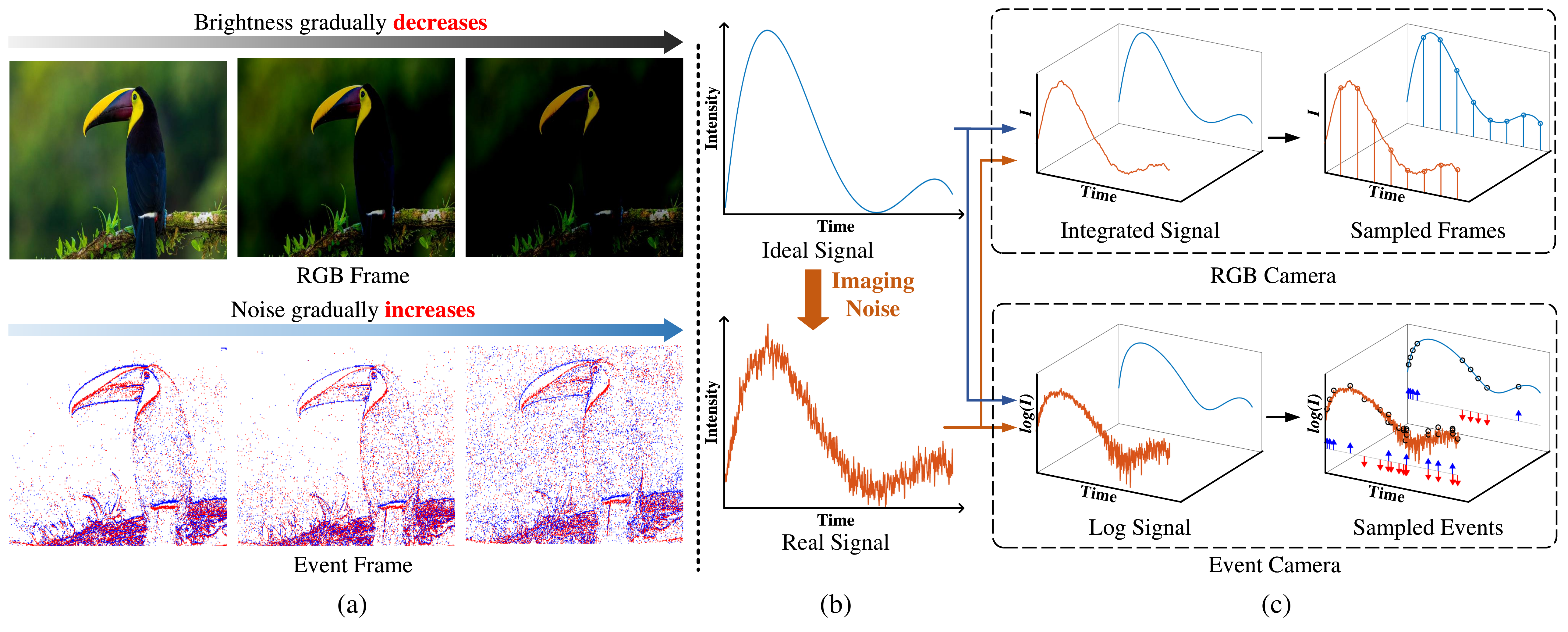}
    \caption{The difference between event and RGB cameras in signal processing. (a) illustrates that the light intensity is inversely correlated to the noise level of captured events, \ie, with the light intensity gradually decreasing, the noise level increases in the event frame\protect\footnotemark[1]. (b)-(c) explain why the event camera generates so much noise in poor lighting conditions. The main reason is that the continual sampling (or differential sampling) in the event camera cannot smooth noisy signals in the integrated sampling manner of a frame-based camera, which makes the event-based denoising task unique and challenging.}
    \label{fig:teaser}
\end{figure*}


\section{Introduction}
\label{section:Intro}

\IEEEPARstart{E}{vent} cameras, such as the Dynamic Vision Sensor (DVS), are novel biologically inspired devices \cite{taverni2018front,wang2019event}. In contrast to traditional frame-based cameras, which capture global scene brightness at a fixed rate, event cameras can asynchronously perceive the environmental brightness change in each pixel and report log-intensity change signals at microsecond resolution \cite{ev-survey,noise-lichtsteiner2008}. These features show great application potential in many fields, such as optical flow estimation \cite{Zhu-RSS-18, flow-zhu2019, flow-pan2020}, high-speed video interpolation \cite{interp-yu2021, interp-timelens, interp-timelens+}, feature tracking/detection \cite{tracking-kueng2016, tracking-eklt, tracking-gallego2017} and simultaneous localization and mapping (SLAM) \cite{slam-weikersdorfer2014, slam-kim2016, slam-evo, zhou2021event, zhang2022unifying, rudnev2021eventhands}.

However, due to its differential imaging mechanism, the event camera is sensitive to various types of noise \cite{noise-lichtsteiner2008, noise-czech2016}. In this paper, we are mainly concerned with background activity (BA) noise \cite{noise-temperature}, which is the main type of noise in event cameras. As shown in Fig.~\ref{fig:teaser} (a), with the overall brightness reduction, the noise level in the event camera output will gradually increase. More specifically, the input signal will be disturbed due to the perturbation of incoming light before the receiving photodiodes and junction leakage current of the imaging circuit, as shown in Fig.~\ref{fig:teaser} (b). In conventional cameras, such noise input will be suppressed to a great extent because of the smoothness of the integration function, thus maintaining good imaging quality. However, in the event camera, the noise is much more obvious due to the continuous differential sampler, and the logarithmic operation will further amplify the noise, leading to the production of BA, as shown in Fig.~\ref{fig:teaser} (c).

Several event denoising datasets \cite{edn-edncnn, edn-eguide, edn-eventzoom} and denoising metrics \cite{edn-truenorth,edn-ynoise,edn-edncnn} have been proposed to date. Based on these, various event denoising algorithms \cite{edn-ynoise, edn-truenorth, edn-seqx, edn-nnb, edn-liu, edn-time_surface, edn-ksvd, edn-knoise, edn-joint, edn-iets, edn-fsae, edn-evflow, edn-eventzoom, edn-eguide, edn-edncnn, wu2020denoising} have been presented and have achieved remarkable progress. However, existing event denoising datasets and denoising metrics still have the following limitations: 1) the scale of existing datasets is small, and the noise diversity is limited and unable to cover authentic working environments of event cameras. Specifically, the events in existing datasets are mainly captured in similar lighting conditions, resulting in small variances in different event sequences, which cannot cover the real noise distribution in practical environments. 2) The existing denoising evaluation metrics are mostly reference evaluation metrics, relying on active pixel sensors (APS) and inertial measurement unit (IMU) information \cite{edn-edncnn} or manual annotation \cite{edn-truenorth, edn-ynoise}. However, APS information is not always available, and its quality cannot be guaranteed, especially in low-light environments. In addition, the microsecond event camera can output millions of events per second, and it is impractical to label each event manually.

To better study the influence of noise on event-based visual cues and enable future research on event denoising, we propose a large-scale event denoising dataset and a nonreference event denoising metric. First, we construct a novel large-scale event denoising dataset, which has three advantages over existing datasets: 1) \textit{Various scenes.} The number of sequences in the E-MLB dataset is 12 times larger than existing datasets. 2) \textit{Varying light conditions.} To better cover the actual lighting conditions in DVS working environments, we collected a large number of event sequences at different times (from day to night). We placed neutral density (ND) filters with fractional transmittances of 1/4, 1/16 and 1/64 in front of the event camera to simulate different light intensities. Thus, four event sequences with different noise levels were obtained for any given scene. 3) \textit{Multiple motion types.} Our dataset contains events generated by objects with different motion types, including translation, rotation, and a combination of both 2D and 3D with perspective changes.

Second, we propose a novel nonreference event denoising metric, termed the event structural ratio (ESR), to reduce the dependence of evaluation metrics on APS information and manual annotation. It has the following advantages: 1) \textit{Effective ranked noise level.} The principle of ESR is to judge the noise level of an event stream. Since each denoising method leads to different noise levels, we can use ESR to evaluate these denoising events and, as a result, distinguish the denoising effect indirectly; 2) \textit{Reflect the intrinsic property of events.} The calculated ESR is not dependent on either the number of events or the projection directions. Therefore, it is an intrinsic property of the events alone. 3) \textit{Easy to calculate.} The only information needed to calculate ESR is the event data, and only basic arithmetic operations are needed.

\footnotetext[1]{The event frame is obtained by accumulating events for each pixel, where red represents positive events and blue represents negative events.}

In summary, the main contributions of this paper are threefold:

\begin{itemize}
    \item{We construct a large-scale event denoising dataset Multi-Level Benchmark (E-MLB) for the first time, which is 12 times larger than the largest existing dataset. Our proposed dataset far exceeds existing datasets in rich real-world scenes and multiple noise levels.}
    \item{We propose the first nonreference event denoising metric, the event structural ratio (ESR), which measures the structural intensity of events without additional information sources such as the APS frame and IMU data. The proposed ESR is easy to calculate and faithfully indicates the noise level of event data under various scenes and lighting conditions.}
    \item{We conduct extensive experiments with 11 state-of-the-art denoising methods on the E-MLB dataset and give the ESR score of each algorithm. We hope that the comparative analysis will contribute to future event denoising research.}
\end{itemize}

The remainder of the paper proceeds as follows. In Section~\ref{section:Related}, we introduce relevant works on event denoising datasets, metrics and algorithms. In Section~\ref{section:Dataset}, we describe the collection details of our E-MLB dataset. Then, we illustrate our proposed event denoising metrics and provide a detailed rigorous mathematical proof in Section~\ref{section:ESR}. In Section~\ref{section:Experiment} and Section~\ref{section:Discussion}, we report the experimental results and give a conclusion, respectively.

\section{Related Works}\label{Related Work}
\label{section:Related}

\subsection{Event Denoising Datasets}

Some denoising datasets have been presented recently to suppress the impact of noise on event cameras. DVSNOISE20 \cite{edn-edncnn} provides 48 event sequences on 16 stationary scenes, which were captured by a \textit{DAVIS 346} mounted in a gimbal restricted to rotation-only movement. It also provides ground-truth labels representing event generation probability by combining the APS and IMU information. ENFS \cite{edn-eventzoom} contains 100 sequences. \textit{DAVIS 346} camera was mounted on top of a table and shot a monitor playing the need-for-speed (NFS) \cite{dataset-nfs} dataset. RGB DAVIS \cite{edn-eguide} provides 20 real event sequences from a \textit{DAVIS 240} camera, including indoor and outdoor scenes, as well as high-resolution frames from a conventional RGB camera. Although these datasets provide a large quantity of realistic noisy data, they were collected under limited lighting conditions; some of them (\eg DVSNOISE20 and ENFS) contain only restricted motion, which cannot cover authentic camera working scenarios.

To solve the lack of ground truth labels, DND21 \cite{edn-low_cost} collected realistic pure noise and pure signal sequences and then synthesized hybrid noisy sequences. Additionally, some simulators, such as ESIM \cite{simulator-esim} and V2E \cite{simulator-v2e}, can be used to generate synthetic DVS events from provided image or video datasets and control noise generation. However, due to the complexity of the actual noise distribution, the above methods cannot reflect real situations.

\subsection{Event-based Denoising Metrics}

Percentage of signal/noise remaining (PSR/PNR) \cite{edn-truenorth} treats the events that fall in the manually generated bounding box as signals, calculating the percentage of remaining events inside (or outside) the bounding box. Noise in real (NIR) \cite{edn-ynoise} and relative plausibility measure of denoising (RPMD) \cite{edn-edncnn} annotate the probability of each event. The former convolves the event stream with a Gaussian kernel, and the latter combines APS and IMU to calculate the probability of event occurrence in each space-time coordinate. In addition, there are some metrics designed on synthetic datasets. Event denoising precision (EDP) \cite{edn-graph} can briefly report the ratio of the total number between the denoised event stream and the original event stream. \cite{edn-low_cost} plots receiver operating characteristic (ROC) curves to compare different event data.

Although the aforementioned metrics can evaluate denoising algorithm performance, some methods rely heavily on synthetic data and this generalization to real event data is still unclear. Others need ground truth data by either manually labeling or introducing additional information sources, which may become invalid in a practical environment where labels are not always available.

\subsection{Event-based Denoising Algorithms}

Statistical methods were the earliest classical approaches for event-based denoising. In \cite{edn-nnb}, outliers are filtered by calculating the density for each event in their local spatial-temporal neighborhood and setting the threshold to judge low-density events. Then, based on this theory, approaches such as \cite{edn-liu, edn-knoise, edn-hash,edn-low_cost,edn-graph} reduce the operating complexity by setting different event storage strategies. Other works, such as \cite{edn-binary, edn-truenorth, edn-ynoise}, introduce additional process stages to eliminate dead pixels or sharpen edges. However, these density statistics methods are difficult to apply across a wide variety of noise and require manually finetuning parameters to deal with different scenarios.

Other algorithm filters conduct event denoising in the context of surface fitting. EV-Gait \cite{edn-evflow} performs local plane optical flow estimation and filters noisy events to smooth the optical flow surface. Afterward, the guided event filter (GEF) \cite{edn-eguide} combines the gradient of active pixel sensor (APS) frames. In contrast, time surface (TS) \cite{edn-time_surface, edn-fsae, edn-iets} transforms events from unit impulses into a representation that is monotonically decreasing with time, which solves the sparsity problem in the local plan fitting process. These fitting methods are well suited for a single moving object but perform poorly in low-light conditions or complex scenarios.

Learning-based methods have been widely used in event-based denoising most recently. For example, a K-SVD method \cite{edn-ksvd} was proposed to extract the sparse features from several noise-free event frames. In \cite{edn-low_cost}, a multilayer perceptron denoising filter (MLPF) was used to calculate the probability of noise event-by-event. In addition, some convolutional neural network (CNN) methods \cite{afshar2020event, edn-edncnn, li2021event, edn-eventzoom} have also been proposed recently. EDnCNN \cite{edn-edncnn} trains a binary classification network using the probability tag of each event, which is estimated by combining APS and IMU information. EDnCNN can classify individual events as signals or noise well but is a time-consuming network. EventZoom \cite{edn-eventzoom} is a high-efficiency U-Net that achieves event denoising in a noise-to-noise fashion.

\section{E-MLB Dataset}
\label{section:Dataset}

In this section, we introduce the collection details of our E-MLB dataset. We first introduce our capture device. Then, the shooting details and photographic accessories used are presented. Finally, the comparison of E-MLB with the existing datasets is given.

\subsubsection*{\bf Event Sensor} 
\begin{figure*}[ht]
    \centering
    \includegraphics[width=\linewidth]{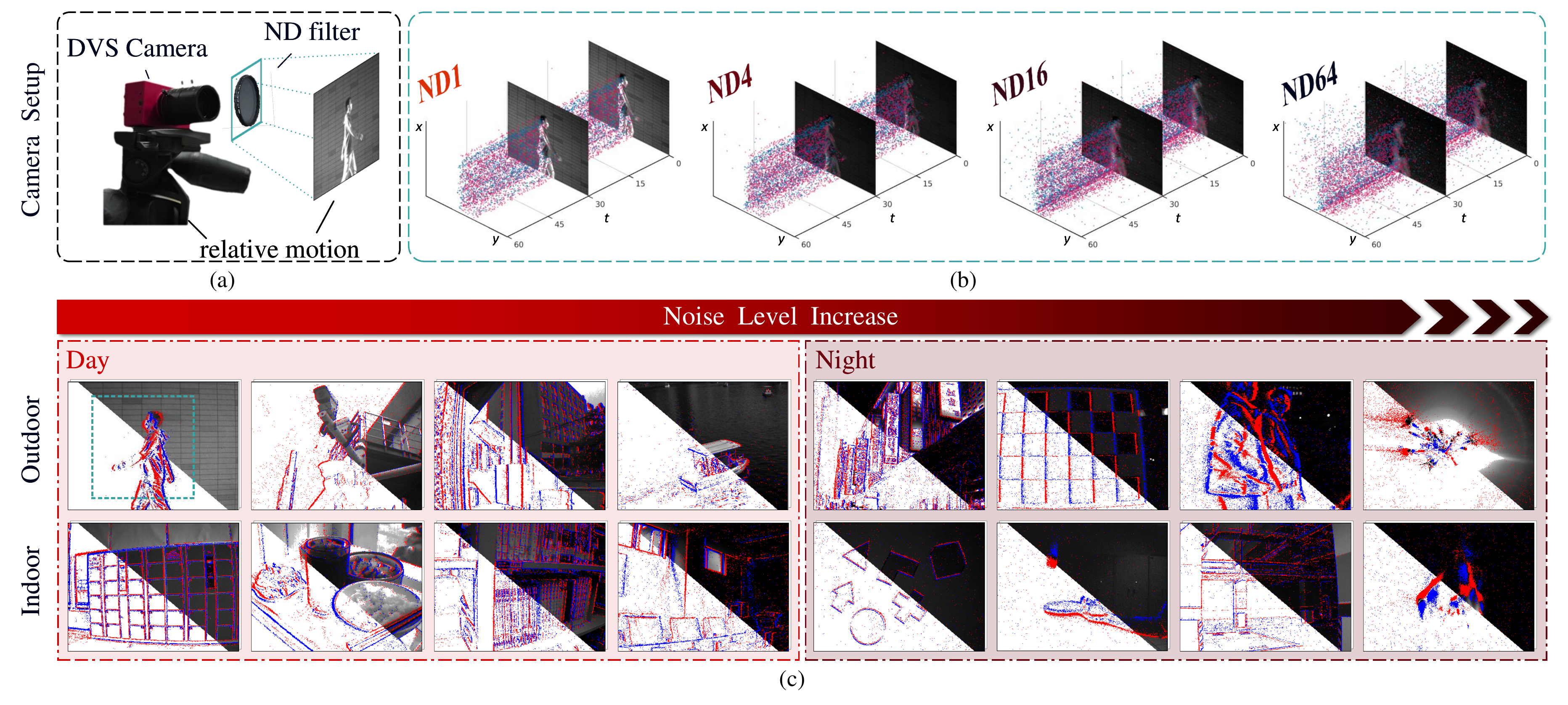}
    \caption{(a) The event camera and ND filters were used for capturing event sequences. (b) The captured event stream with different ND filters\protect\footnotemark[2]. The noise level gradually increases with the amount of light entering the lens reduction. (c) Examples of event sequences in the E-MLB from daytime to night. In each square, the lower-left is the converted event frame, and the upper-right is the hybrid image, including the event and APS frame.}
    \label{fig:Dataset}
\end{figure*}

The type of event camera we used was a \textit{DAVIS346}, which can simultaneously output a spatially aligned event stream ($120$ dB) and intensity images ($56$ dB) with a resolution of $346 \times 260$. In addition, to simulate diﬀerent lighting conditions, we placed three neutral density filters (ND filters) with different transmittance in front of the lens, as shown in Fig.~\ref{fig:Dataset} (a).

\subsubsection*{\bf Collection Details} Benefiting from the high dynamic property, the DVS is widely used in extreme light conditions, such as low-light and overexposed conditions \cite{recon-e2vid,recon-hdr}. However, the noise level output by DVS gradually increases as the light intensity increases/decreases, as shown in Fig.~\ref{fig:teaser}. To better analyze the relationship between noise level and light intensity, we place the ND filter to simulate the different light conditions, as shown in Fig.~\ref{fig:Dataset} (a). For each scene, we first capture the original scene in the natural light condition. Then, we add the ND filter in front of the DVS and repeat the capture process. In this paper, we use three kinds of ND filters with different transmittance (1/4, 1/16, and 1/64), which are denoted as ND4, ND16, and ND64, respectively. The captured samples are shown in Fig.~\ref{fig:Dataset} (b). For each light condition, we repeatedly shoot the scene 3 times. The simulated light intensity and diversity are highly dependent on the original natural light intensity. Thus, to further increase the light diversity, we change the capture time from day to night to guarantee the diversity of natural lighting conditions.

\begin{table*}[h]
    \begin{center}
        \caption{The comparison of our proposed E-MLB with existing event denoising dataset.}
        \label{tab:dataset comparison}
        \resizebox{\textwidth}{!}{
            \begin{tabular}{lcccccccccc}
                \toprule
                Datasets                     & Camera             & Resolution              & APS  & IMU*       & Scenes       & Sequences     & Capture$/s$   & DoF  & Noise Level & \\
                \midrule
                DVSNOISE20 \cite{edn-edncnn} & \textit{DAVIS 346} & \textit{$346\times260$} & Gray & \checkmark & 16           & 48            & 807           & Cam. & -           & \\
                RGBDAVIS \cite{edn-eguide}   & \textit{DAVIS 240} & \textit{$190\times180$} & RGB  & -          & 20           & 20            & 122           & All. & -           & \\
                ENFS \cite{edn-eventzoom}    & \textit{DAVIS 346} & \textit{$224\times125$} & -    & -          & 1            & 100           & 4238          & Obj. & -           & \\
                DND-21 \cite{edn-low_cost}   & \textit{DAVIS 346} & \textit{$346\times260$} & -    & -          & -            & 8             & -             & All. & -           & \\
                E-MLB                        & \textit{DAVIS 346} & \textit{$346\times260$} & Gray & \checkmark & \textbf{100} & \textbf{1200} & \textbf{7300} & All. & \textbf{4}  & \\
                \bottomrule
            \end{tabular}
        }
        \begin{tablenotes}[flushleft]\footnotesize\smallskip
            \item{*} Inertial Measurement Unit
        \end{tablenotes}
    \end{center}
\end{table*}

In addition to changing the light intensity, we also change the shooting scene to guarantee the diversity of the content of the captured event sequence. In this paper, we select 100 scenes, including both indoor and outdoor scenes and diverse motion types (translation, rotation, and combination of both in 2D and 3D with perspective change). In addition, we provide the corresponding APS frame and IMU data for each captured event sequence. It should be noted that the APS quality will decrease as light intensity decreases. Considering that the event camera has a superior high dynamic range, we also include some special sequences that create more challenges for denoisers, such as extremely low light scenes (with high background activity and blurred edges), special weather conditions (rainy and snowy days), and high-speed objects. Some example sequences can be found in Fig.~\ref{fig:Dataset} (c). A comparison of our E-MLB with the existing event denoising dataset is reported in Tab.~\ref{tab:dataset comparison}.

\footnotetext[2]{For consistency, ND1 is used to represent the data captured without any ND filters}

\section{Event Structural Ratio}
\label{section:ESR}

In \cref{ESR:preliminaries}, we review the working principle of event cameras and the event contrast measurement, in which event contrast is the main inspiration of our denoising metric. In \cref{ESR:definition}, we introduce our proposed event structure ratio, and the relevant derivation and proof can be found in \cref{ESR:proof}. Evaluations on ESR are conducted in \cref{ESR:experiment}, including both synthetic and real experiments, which demonstrate that our proposed ESR is a good denoising indicator.

\subsection{Preliminaries\label{ESR:preliminaries}}

\textbf{Working Principle:} In event cameras, each pixel works asynchronously and will trigger an event $e_k := (\bm{x}_k, t_k, p_k)$ when its logarithmic brightness change reaches the predefined contrast threshold $c$, which can be defined as:
\begin{equation}
    \Delta L \doteq L(\bm{x}_k, t_k) - L(\bm{x}_k, t_k - \Delta{t_k}) = c \cdot p_k \label{eqs:principle}
\end{equation}
where $\bm{x}_k \coloneqq (x_k, y_k)$ is the pixel position of the $k$-th event. $t_k$ is the timestamp, and $\Delta{t_k}$ is the time interval since pixel $(x_k, y_k)$ last reaches the threshold. $p_k\in\{-1,+1\}$ is the polarity, representing the decrease and increase in brightness, respectively. $L(\bm{x}_k, t_k) \coloneqq \log I(\bm{x}_k, t_k)$ denotes the log intensity.

The difference between log intensity in a duration of $t$ can be obtained by integrating the sequences of events \cite{lin2022autofocus}:
\begin{equation}
    L(\bm{x}, t) - L(\bm{x}, 0) \doteq c \cdot \int_0^t \sum_k e_k(\bm{x}, \tau) d\tau \label{eqs:interg}
\end{equation}
where $e_k(\bm{x}, t)$ can be described by using Dirac function $\delta(\cdot)$:
\begin{equation}
    e_k(\bm{x}, t) = p_k \cdot \delta(\bm{x}-\bm{x}_k, t-t_k)
\end{equation}

\begin{figure}
    \centering
    \includegraphics[width=\linewidth]{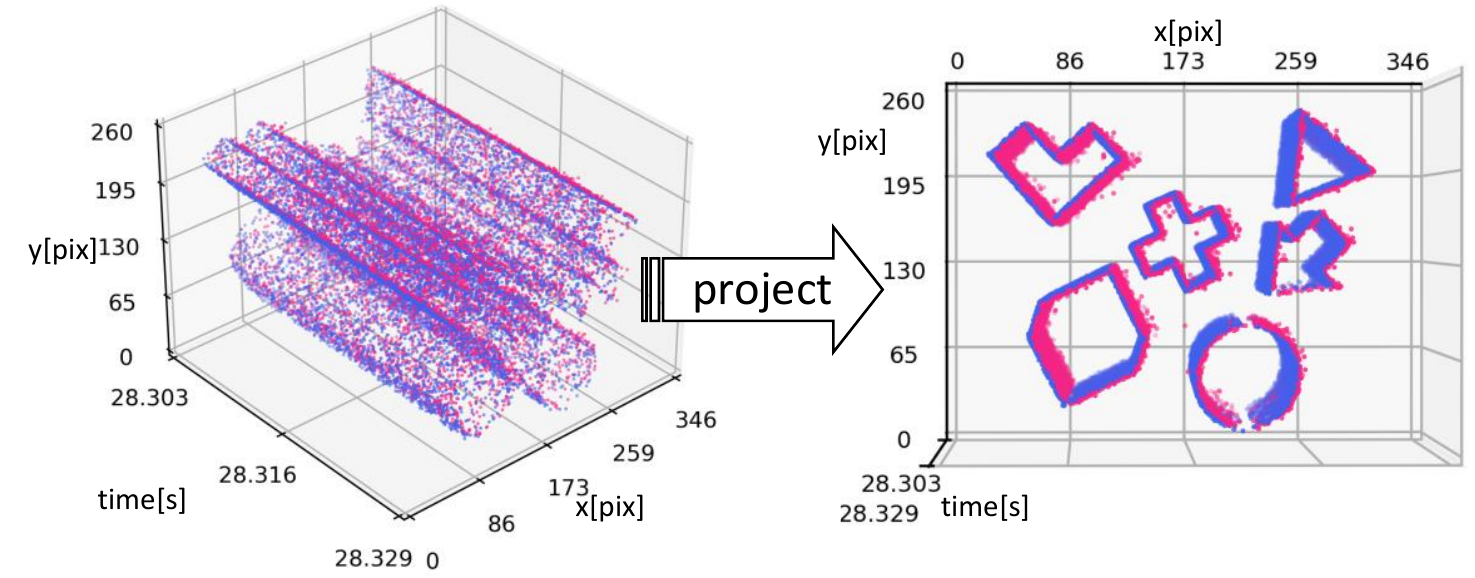}
    \caption{Since the event camera is responsive to edges, we will obtain an image of the objects’ edges after projecting the events along the trajectories to the 2D dimension, which helps us to analyze its statistical characteristics more easily.}
    \label{fig:warpping}
\end{figure}

\textbf{Event Contrast:} Since event cameras are highly responsive to the moving edges of an object \cite{ev-cmax}, a set of events will occur on the edge trajectories as long as relative movement occurs between the camera and objects. In contrast, given a set of events $\{e_k\}_{N}$, we can project (warp) these events to a reference time $t_{ref}$ along these trajectories by a warping function $W$:
\begin{equation}
    e_k \coloneqq (\bm{x}_k,t_k,p_k)\overset{W}{\mapsto} e_k' \coloneqq (\bm{x}_k',t_{ref},p_k)
\end{equation}
After projecting, we obtain an accumulated 2D histogram, also known as an image of warped events (IWE) \cite{gallego2019focus}:
\begin{equation}
    \text{IWE}(\bm{x})=\sum_{k=1}^N b_k\delta(\bm{x}-\bm{x}_k')
\end{equation}
where $b_k$ is the weight of the summation of $e_k$. Here, we set $b_k=1$ to facilitate the subsequent derivations. Usually, the warping function can be modeled as linear motion (optic flow), rotational motion, 4-DOF motion and so on \cite{chen2022progressivemotionseg}. If we correctly model the warping function and estimate accurate parameters, the IWE will form an edge-like image. Taking \cref{fig:warpping} as an example, for some simple shapes performing translation motion relative to the camera, we can project events along the translation direction to obtain a clear and sharp edge-like IWE.

Because edge strength is directly related to image contrast \cite{gallego2019focus}, we can use IWE to measure scene contrast. Here, we use an image-based contrast metric named the total sum of squares (TSS):
\begin{equation}
    \text{TSS}=\sum_{\bm{x}}\text{IWE}^2(\bm{x}),
\end{equation}
where the summation is carried over all the pixels. The area of spatial support $L$ (the total number of pixels that output events) can be defined as:
\begin{equation}
    L \coloneqq \sum_{\text{IWE}(\bm{x})>0}1,
\end{equation}
TSS and $L$ are inversely correlated most of the time. Given a number of events, the more aggregated the events are in IWE, the less spatial support $L$ the event image has. In other words, the event contrast will decrease when the data are influenced by noise, which we believe is an important clue to judging the impact of noise.

However, TSS and other contrast metrics are highly dependent on the number of events, and they cannot be directly used as event denoising metrics. Taking $TSS$ as an example, it will always assign the highest score for the denoising method that outputs the maximum number of events. In practice, we cannot guarantee that the different denoising methods keep the same number of events.

\subsection{Definition of ESR\label{ESR:definition}}

To address the above issues, we extract an invariant from the TSS, which is called the normalized TSS (NTSS):
\begin{equation}
    \text{NTSS} \coloneqq \sum_{i=1}^K\frac{n_i(n_i-1)}{N(N-1)}
\end{equation}
where $K$ is the total number of pixels in the IWE. $N$ is the total number of events, and $n_i$ is the sum of all events that occur on pixel $(x_i, y_i)$. NTSS is used to represent the relative contrast of the scene regardless of the number of events. Nevertheless, due to the intrinsic deficiency of the contrast metric, the NTSS tends to assign a higher score to the method that performs overdenoising. An extreme case is that if only one event remains, the calculated NTSS will reach the upper bound and fail to faithfully represent the noise situation. Therefore, we add a penalty coefficient before NTSS, which is defined as:
\begin{equation}
    L_N \coloneqq K-\sum_{i=1}^K(1-\frac{M}{N})^{n_i}
\end{equation}
where $L_N$ is the number of nonzero pixels (or the area of spatial support) in the IWE. $M$ is the reference number of events used for interpolation, which is fixed during the entire evaluation process. In this way, the normalized contrast of any N events can be interpolated to that of fixed M events. Based on the invariant representation of scene contrast NTSS and penalty coefficient $L_N$, we can finally define the proposed ESR as:
\begin{equation}
    \text{ESR}:=\sqrt{\text{NTSS}\cdot L_N},
\end{equation}

\subsection{Proof of NTSS and $L_N$}\label{ESR:proof}

For small duration $\Delta t$, the probability of a given number of events follows the Poisson distribution \cite{edn-knoise}:
\begin{equation}
    P(N_{\bm{x}}(t)=m)=e^{-\lambda_{\bm{x}}t}\frac{(\lambda_{\bm{x}}t)^m}{m!}
\end{equation}
where $P(N_{\bm{x}}(t)=m)$ is the probability of $m$ events occurring. Event rate $\lambda_{\bm{x}}$ is the rate of triggering events at pixel per unit time \cite{lin2022autofocus}, which can be derived from \cref{eqs:interg} as:
\begin{equation}
    \lambda_{\bm{x}} \coloneqq \frac{1}{\Delta t} \cdot \frac{L(\bm{x}, t) - L(\bm{x}, 0)}{c} = \frac{\int_0^t \sum_k e_k(\bm{x}, \tau) d\tau}{\Delta t}
\end{equation}
then we can obtain the uniform event rate as:
\begin{equation}
    p_{\bm{x}} = \frac{\lambda_{\bm{x}}}{\sum_{\bm{x}}\lambda_{\bm{x}}},
\end{equation}
where $p_{\bm{x}}$ represents the relative portion of event rate $\lambda_{x,y}$ and sums to 1. Given the sum of the number of events, the joint probability distribution of the number of events in different pixels follows a multinomial distribution, provided the number of events in each pixel follows the Poisson distribution and all the pixels are independent. Therefore, events can be viewed as drawn from a multinomial distribution provided that the number of events is fixed. Let the total number of pixels of the event image be $K$, and use flattened index $i\in\{1,2,...K\}=(x_i,y_i)$ to represent different pixels for simplicity of notation; then we have:
\begin{align}
    (n_1,...n_K)|Y & =N\sim Multinomial(N,(p_1,...p_K)),         \\
    n_i|Y          & =N\sim Binomial(N,p_i)\label{eqs:binomial}, \\
    Y              & =\sum_{i=1}^Kn_i,                           \\
    n_i            & =N_{\bm{x}_i}(t).
\end{align}

After deriving the uniform event rate $p_{\bm{x}}$, we can further derive the NTSS and $L_N$. The expectation of TSS is:
\begin{equation}
    \E[\text{TSS}|Y=N]=\sum_{i=1}^K\E[n_i^2|Y=N].
\end{equation}
Because the distribution of $N_i(t)$ conditioned on $M(t)$ is binomial, we introduce
\begin{equation}
    f(x,p)=(px+(1-p))^N,
\end{equation}
then from \cref{eqs:binomial}, we have:
\begin{align}
    \E[n_i^2|Y=N] & =\sum_{k=1}^Nk^2\binom{N}{k}p^k(1-p)^{N-k}             \\
                  & =(x\frac{\partial}{\partial x})^2\circ f(x,p_i)|_{x=1}
    =Np_i+N(N-1)p_i^2.
\end{align}
so for TSS, there is:
\begin{align}
    \E[TSS|Y=N] & =\sum_{i=1}^KNp_i+N(N-1)p_i^2 \\
                & =N+N(N-1)\sum_{i=1}^Kp_i^2.
\end{align}
$\sum_{i=1}^Kp_i^2$ is an inherent property of the scene and is invariant with respect to the number of events $N$. In effect, it can be estimated by:
\begin{align}
    \sum_{i=1}^Kp_i^2\approx\sum_{i=1}^K\frac{n_i(n_i-1)}{N(N-1)}.
\end{align}
$\sum_{i=1}^Kp_i^2$ can be viewed as the normalized TSS, and its estimation is denoted as NTSS:
\begin{equation}
    NTSS:=\sum_{i=1}^K\frac{n_i(n_i-1)}{N(N-1)}.
\end{equation}

The expectation of L is:
\begin{align}
    \E[L|Y=N] & =\E[\sum_{i=1}^K1_{n_i>0}|Y=N]=\sum_{i=1}^KP(n_i>0|Y=N) \\
              & =K-\sum_{i=1}^KP(n_i=0|Y=N)
    \approx K-\sum_{i=1}^Ke^{-Np_i}.\label{eqn:Lapprox1}
\end{align}
There is no simple scene invariant from the expression because $N$ and $p_i$ are tightly coupled; however, by introducing a new random variable $\alpha^{n_i}$, we can interpolate the resultant $L$ to any given number of $M$ as if it were calculated by exactly $M$ events. The expectation of this new random variable is:
\begin{align}
    \E[\alpha^{n_i}|Y=N] & =\sum_{k=1}^N\alpha^kp_i^k(1-p_i)^{N-k}\binom{N}{k}                 \\
                         & =(1+(\alpha-1)p_i)^N\approx e^{(\alpha-1)Np_i}.\label{eqn:Lapprox2}
\end{align}
Thus, by setting $(\alpha-1)N=-M$, or equivalently $\alpha=1-\frac{M}{N}$, we can interpolate any $L$ when $Y=M$ from $N$ events, defined as:
\begin{equation}
    L_N:=K-\sum_{i=1}^K(1-\frac{M}{N})^{n_i}.
\end{equation}

\begin{figure}
    \centering
    \includegraphics[width=\linewidth]{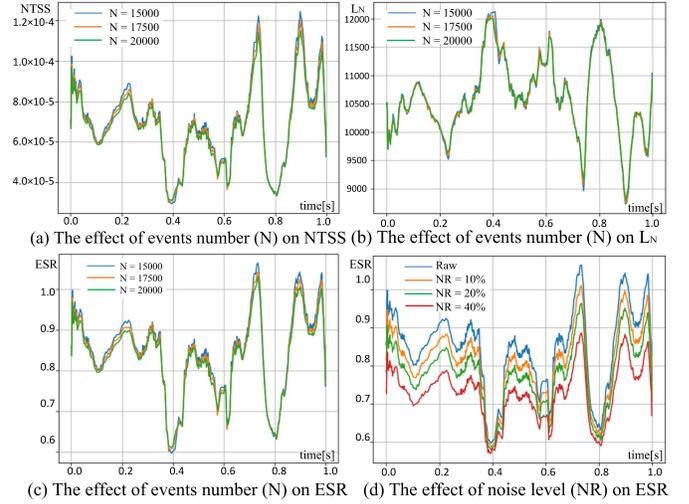}
    \caption{The effect of event number and noise level on ESR. The NTSS and $L_N$ are robust to the number of events, as in (a) and (b), which results in the obtained ESR also being robust to the number of events, as in (c). (d) shows that the proposed ESR is inversely correlated to the noise level, and a higher noise level corresponds to a lower ESR.}
    \label{fig:Fig3}
\end{figure}

\subsection{Experimental Verification\label{ESR:experiment}}
To verify that the proposed NTSS, $L_N$ and ESR are independent of the number of events, we conduct experiments on real-world event sequences. We conduct three experiments with N = 15,000, 17,500, and 20,000. $M$ is set to $15,000$ in all three experiments. Events in the whole sequence are split into packets of events with equal sizes of $N$. Then, we compute the NTSS, $L_N$, and ESR values for each event packet with predefined parameters and draw the NTSS, $L_N$, and ESR curves of the entire sequence. As shown in \cref{fig:Fig3} (a-b), when the number $N$ changes from 15,000 to 20,000, the NTSS and $L_N$ curves are very close, which verifies their independence from the event number $N$. As a result, the ESR curve is also independent of $N$, as shown in \cref{fig:Fig3} (c). Then, we test the relationship between the ESR and noise level. We manually add random noise (the noise ratio is set to 10\%, 20\%, and 40\%) to the original sequence and calculate the corresponding ESR curve. As shown in \cref{fig:Fig3} (d), the noisy ESR curves have the same shape as the original curve, and the noisier the ESR curve is, the lower the ESR value, which validates that it can indicate the noise level and can be used as an event-based denoising evaluation metric.

\begin{figure}
    \centering
    \includegraphics[width=0.95\linewidth]{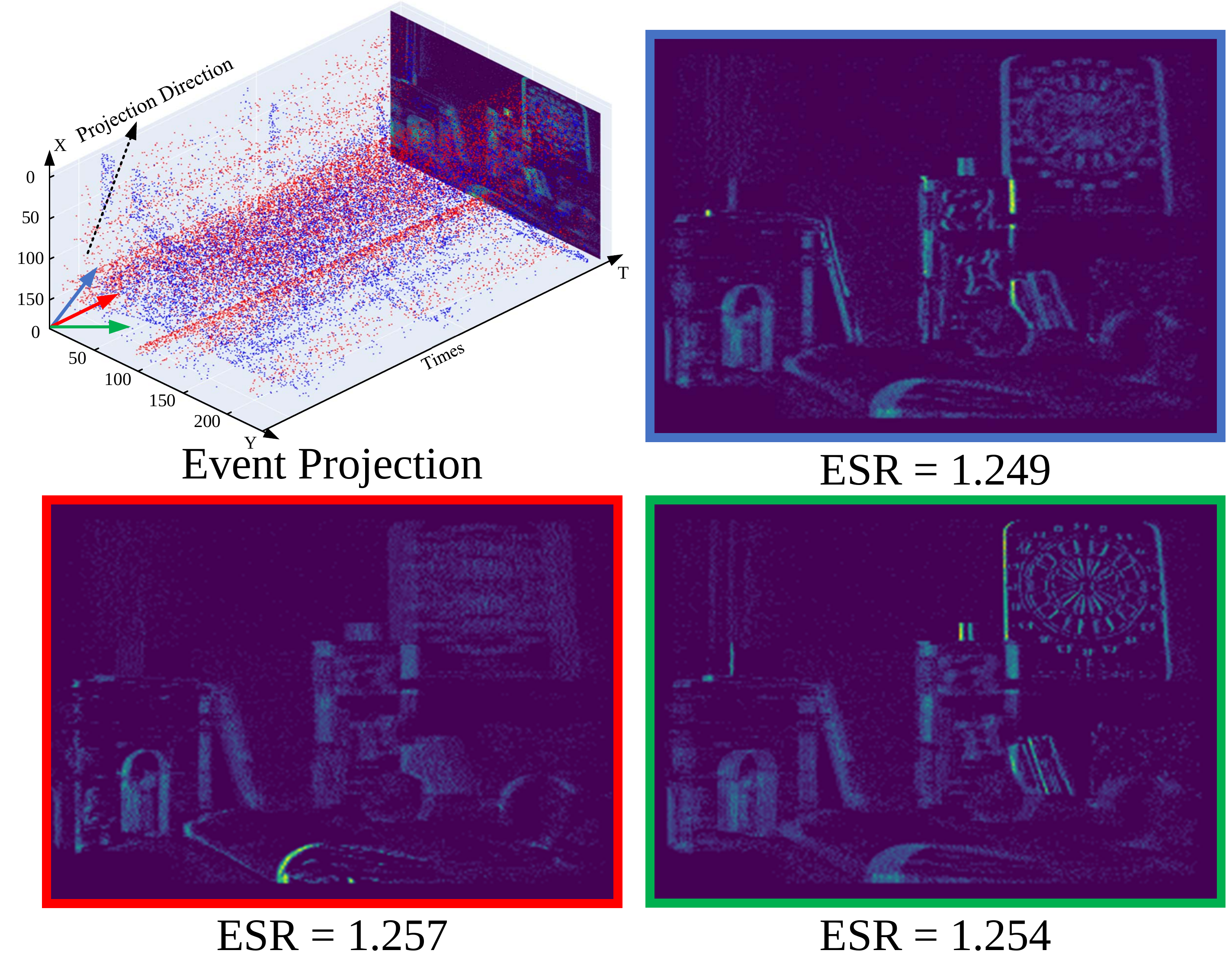}
    \caption{The effect of projection directions on ESR. Although events are projected along different directions, the ESR values are relatively close.}
    \label{fig:Fig4-ProjectionManner}
\end{figure}

The proposed ESR is calculated in the event frame to simplify the calculation, whereas the existing algorithms adopt different projection methods during the process. For example, EventZoom uses projections along the time axis, and GEF uses projections along the motion axis, which leads to a change in the event distribution after denoising. Therefore, we need to test the influence of different projection methods on the ESR value to verify its robustness on different algorithms. As shown in \cref{fig:Fig4-ProjectionManner}, we calculate the ESR value of the same event packet in different projection directions; the resultant ESR values are quite close, so the proposed ESR is also invariant to the change in projection direction. In conclusion, although the calculation is performed in the frame space, the resultant ESR is independent of the number of events and the projection method only represents the event quality, which is thus, an intrinsic property of events.

\section{Experimental Results}
\label{section:Experiment}

In this section, we first provide the mean ESR ({\bf MESR}) score of each representative denoiser in both our E-MLB and other existing datasets and present some typical visualization results. Then, a comparison of ESR with another denoising metric is given, which proves the superiority of ESR.

\begin{table*}[]
    \small
    \caption{The mean ESR (MESR) results of different denoising methods on both E-MLB dataset and public available event denoising datasets. We mark the \textbf{{best}} and \underline{{second best}}.}
    \label{tab:ESR results}
    \resizebox{\textwidth}{!}{
        \begin{tabular}{lcccccccccccccccccc}
            \toprule
            \multicolumn{1}{c}{}           & \multicolumn{4}{c}{E-MLB (Daytime)} &                      & \multicolumn{4}{c}{E-MLB (Night)} &                      & \multicolumn{2}{c}{RGB DAVIS} &                      & DVS NOISE20          &                      & ENFS                 &  & DND21                                                                                                                     \\ \cline{2-5} \cline{7-10} \cline{12-13} \cline{15-15} \cline{17-17} \cline{19-19}
            \multicolumn{1}{c}{}           & ND1                                 & ND4                  & ND16                              & ND64                 &                               & ND1                  & ND4                  & ND16                 & ND64                 &  & Indoor               & Outdoor              &  & -                    &  & -                    &  & -                    \\ \cline{1-19}
            Raw                            & 0.821                               & 0.824                & 0.815                             & 0.786                &                               & 0.890                & 0.824                & 0.786                & 0.768                &  & 0.905                & 0.776                &  & 0.524                &  & 0.843                &  & 0.869                \\
            BAF \cite{edn-nnb}             & 0.861                               & 0.869                & 0.876                             & 0.890                &                               & 0.946                & 0.973                & 0.992                & 0.942                &  & 0.943                & 0.891                &  & 0.600                &  & 1.119                &  & 0.920                \\
            KNoise \cite{edn-knoise}       & 0.846                               & 0.837                & 0.830                             & 0.807                &                               & 0.954                & 0.956                & 0.871                & 0.817                &  & 0.934                & 0.895                &  & 0.550                &  & 0.945                &  & 0.887                \\
            DWF \cite{edn-low_cost}        & 0.878                               & 0.876                & 0.866                             & 0.865                &                               & 0.923                & 0.962                & 0.988                & 0.932                &  & 0.923                & 0.890                &  & 0.458                &  & 1.108                &  & 0.905                \\
            EvFlow \cite{edn-evflow}       & 0.848                               & 0.878                & 0.868                             & 0.833                &                               & 0.969                & 0.983                & 0.889                & 0.797                &  & 0.829                & 1.061                &  & 0.667                &  & 1.131                &  & {\underline {1.006}} \\
            YNoise \cite{edn-ynoise}       & 0.866                               & 0.863                & 0.857                             & 0.821                &                               & 1.009                & 0.943                & 0.875                & 0.792                &  & 0.825                & 1.077                &  & 0.654                &  & 1.178                &  & 0.966                \\
            TS \cite{edn-time_surface}     & 0.877                               & 0.887                & 0.870                             & 0.837                &                               & {\underline {1.033}} & 0.944                & 0.886                & 0.797                &  & 0.837                & {\underline {1.120}} &  & 0.745                &  & {\underline {1.241}} &  & 0.985                \\
            IETS \cite{edn-iets}           & 0.772                               & 0.785                & 0.777                             & 0.753                &                               & 0.950                & 0.823                & 0.804                & 0.711                &  & 0.762                & 0.988                &  & 0.733                &  & 0.982                &  & 0.900                \\
            GEF \cite{edn-eguide}          & {\textbf {1.051}}                   & {\underline {0.938}} & 0.935                             & 0.927                &                               & 1.027                & 0.955                & 0.946                & 0.935                &  & {\textbf {1.031}}    & 0.986                &  & {\underline {1.010}} &  & 1.152                &  & 0.932                \\
            MLPF \cite{edn-low_cost}       & 0.851                               & 0.855                & 0.846                             & 0.840                &                               & 0.926                & 0.928                & 0.910                & 0.906                &  & {\underline {0.983}} & 0.932                &  & {\textbf {1.041}}    &  & 1.132                &  & 0.944                \\
            EDnCNN \cite{edn-edncnn}       & 0.887                               & 0.908                & {\underline {0.903}}              & {\underline {0.912}} &                               & 1.001                & {\textbf {1.024}}    & {\textbf {1.079}}    & {\textbf {1.086}}    &  & 0.982                & 1.014                &  & 0.862                &  & 1.232                &  & 0.977                \\
            EventZoom \cite{edn-eventzoom} & {\underline {0.996}}                & {\textbf {0.988}}    & {\textbf {0.996}}                 & {\textbf {0.970}}    &                               & {\textbf {1.055}}    & {\underline {1.007}} & {\underline {1.010}} & {\underline {0.988}} &  & 0.930                & {\textbf {1.135}}    &  & 0.899                &  & {\textbf {1.417}}    &  & {\textbf {1.059}}    \\ \bottomrule
        \end{tabular}
    }
\end{table*}

\subsection{Event Denoising Baselines}
We select the 11 most representative event denoising methods for comparison:

\begin{itemize}
    \item {\bf BAF \cite{edn-nnb}, KNoise \cite{edn-knoise} \& \bf DWF \cite{edn-low_cost}} follow the same denoising theory. The background activity filter (BAF) counts the density of each incoming event in its eight neighborhood pixels within a time interval and filters out noise events according to a predetermined threshold. KNoise improves on this basis by allocating two blocks of memory to store the latest events of rows and columns, which gains the advantage of $O(N)$ space complexity. The double window filter (DWF) further reduces the memory footprint by using a first-in-first-out (FIFO) queue, which stores only a few recent events and determines whether to insert a new event into this queue by comparing it with in-queue events.
    \item {\bf TS \cite{edn-time_surface} \& IETS \cite{edn-iets}} convert a sparse event stream into a dense representation. First, the time surface (TS) converts the Dirac function of time into a logarithmic decay representation, in which case the effective events form a regular manifold called the time surface. Then, it eliminates events that destroy the smoothness of the surface. The inceptive event time surface (IETS) introduces predefined time thresholds to eliminate redundant events within the same edge.
    \item {\bf EvFlow \cite{edn-evflow}} calculates the gradient by local plane fitting to attain optical flow and then achieves event denoising by filtering all the events with abnormal flow values.
    \item {\bf YNoise \cite{edn-ynoise}} calculates the density of each incoming event in its spatiotemporal domain and then achieves event denoising by passing events with high density.
    \item {\bf MLPF \cite{edn-low_cost}} is a kind of multilayer perceptron (MLP) method with a single hidden layer, which is trained by adding simulated noise events in the noise-free sequences.
    \item {\bf EDnCNN \cite{edn-edncnn}} is a convolutional neural network. The probability of an event can be calculated by fusing APS and IMU data, which are used as the labels for each training event.
    \item {\bf GEF \cite{edn-eguide}} provides two denoising modes. In the frame-guide mode, the guided event filter (GEF) extracts mutual structures between the event frame (project along optical flow) and the gradient of the APS image (by Sobel operator), then deletes unreasonable events and reallocates back to spatiotemporal space. When the APS quality is not high, GEF changes to self-guide mode, aligning two adjacent event frames and erasing inconsistent events.
    \item {\bf EventZoom \cite{edn-eventzoom}} follows a noise-to-noise fashion that utilizes paired noisy event sequences to train a U-net and performs event reconstruction guidance using good quality videos on the network branch.
\end{itemize}

\subsubsection*{\bf Experimental Details} All sequences in the E-MLB dataset are tested with the above denoising algorithms. To ensure a fair comparison, we manually fine-tune the adjustable parameters of all methods in each sequence. It should be noted that EDnCNN trained on our dataset performs inferior to the pretrained EDnCNN. The reason is that EDnCNN is highly dependent on its exclusive event noise probability labels, which are restricted to stationary scenes with rotation-only camera motion; otherwise, it will be trained with incorrect training data, and our dataset does not strictly meet this requirement. Therefore, we choose a pretrained network on their DVSNOISE20 dataset and then fine-tune it on our rotation-only sequences. In terms of GEF, we set the frame-guide model in ND1 sequences while changing to self-guided in ND4, ND16 and ND64 sequences. As mentioned in Section 3, in ND1 sequences frame-guide performs better because of the high-quality frames. However, in ND4 to ND64 frames, the quality falls and the self-guided mode can provide more reasonable denoising results. Considering that we do not provide similar paired noise sequences as in the ENFS dataset, we only trained EventZoom on its ENFS dataset sequences.

To calculate MESR, we slice the event sequence $\mathcal{E} \coloneqq \{e_k\}$ consecutively along the time, which can make the set of nonoverlapping event groups $\{ \{e_k\}^1, \{e_k\}^2, \dots, \{e_k\}^G \subseteq \mathcal{E}\}$, where $G$ is the number of event groups. Each group is a subset that belongs to the original sequence. In the experiment, we specified that each event group contains 30,000 events; therefore, we chose $M=20,000$ and $N=30,000$ for all sequences for a fair comparison.

\subsection{Experimental Results}

\subsubsection*{\bf Quantitative Evaluation} The mean ESR (MESR) results are reported in \cref{tab:ESR results}. As shown in the first row, the MESR score of the E-MLB dataset decreases as the noise level increases (from ND1 to ND64), which again verifies the inverse correlation between the ESR value and noise level. The only exception is that the MESR score of ND1 is slightly lower than that of ND4 in the daytime sequences. We also provide MESR scores in some other event-based denoising datasets, i.e., RGB DAVIS, DVSNOISE20, ENFS and DND21. Their ESR results are similar to those sequences in our daytime E-MLB dataset because they were all captured under normal light conditions.

For the different denoising methods, it is clear that almost all the denoised sequences report better ESR scores compared to the raw sequences, especially in the higher score improvement in the night sequences. Overall, we can determine that BAF, Knoise and DWF receive approximate ESR scoring results as they follow a similar denoising principle. Considering that IETS eliminates a large number of effective signals, it reports the poorest score. GEF outperforms other denoising methods when the APS quality is good, namely, in ND1 sequences of E-MLB and other datasets that provide related frames. Nevertheless, the denoising score drops to the second tier when GEF enters the self-guided mode. EventZoom reports the highest MESR score in almost all normal light sequences, \eg, E-MLB in the daytime, while EDnCNN presents the best performance when the noise level is higher, as shown in the ND4 to ND64 columns at night E-MLB.

\begin{figure*}
    \centering
    \includegraphics[width=0.98\linewidth]{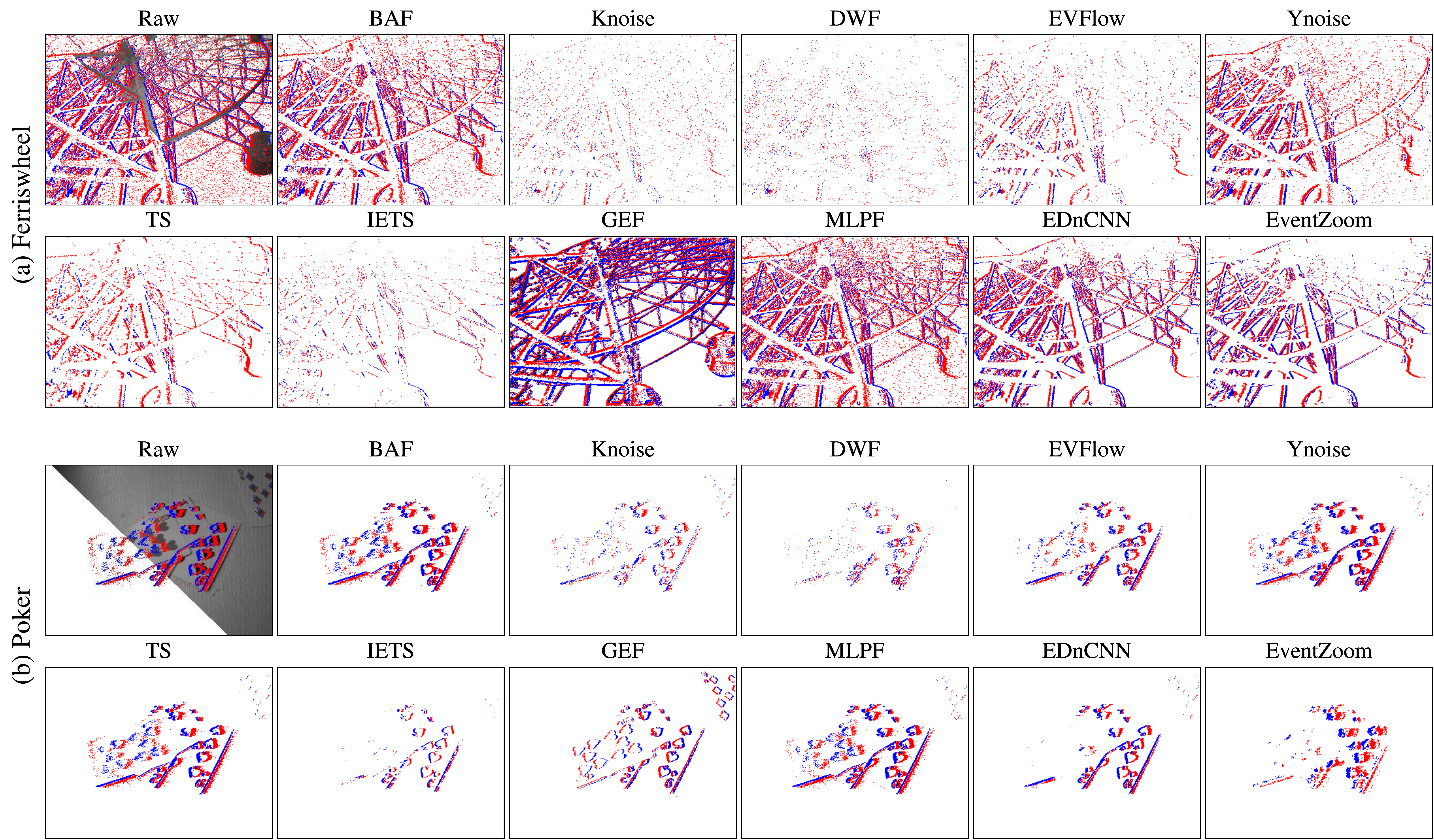}
    \caption{The visual comparison from different denoising algorithms in some representative daytime sequences, including (a) a static object shoot against strong sunlight, in which case a lot of single polarity noise will be generated, and (b) multiple fast-moving objects in a noise-free environment, which is challenging for speed-sensitive denoisers.}
    \label{fig:CrossSceneDaytime}
\end{figure*}

\begin{figure*}
    \centering
    \includegraphics[width=0.98\linewidth]{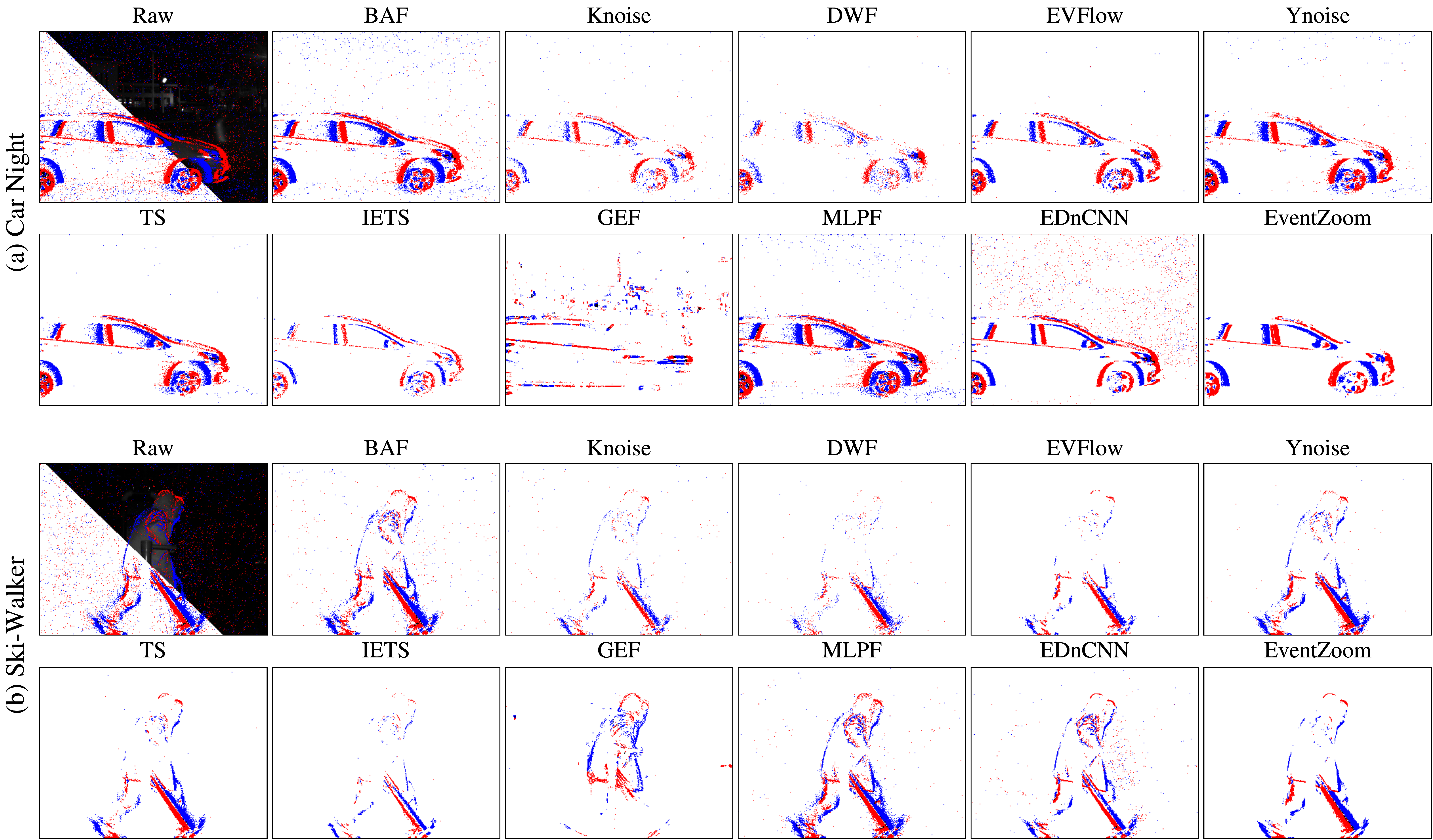}
    \caption{The visual comparison from different denoising algorithms in some representative night sequences, including (a) a vehicle under a street light and (b) nonrigid body motion. Note that in night sequences, we have no choice but to increase exposure times as much as possible to acquire visible frames, which creates some inevitable problems such as smear or blur.}
    \label{fig:CrossSceneNight}
\end{figure*}

It is also worth noting that our ESR still works effectively for algorithms that can generate new possible events (such as EventZoom and GEF). However, the other existing event-based denoising metrics almost fail to evaluate such self-generated events from denoisers, providing lower scores despite good denoising performance.

\begin{figure*}
    \centering
    \includegraphics[width=0.95\linewidth]{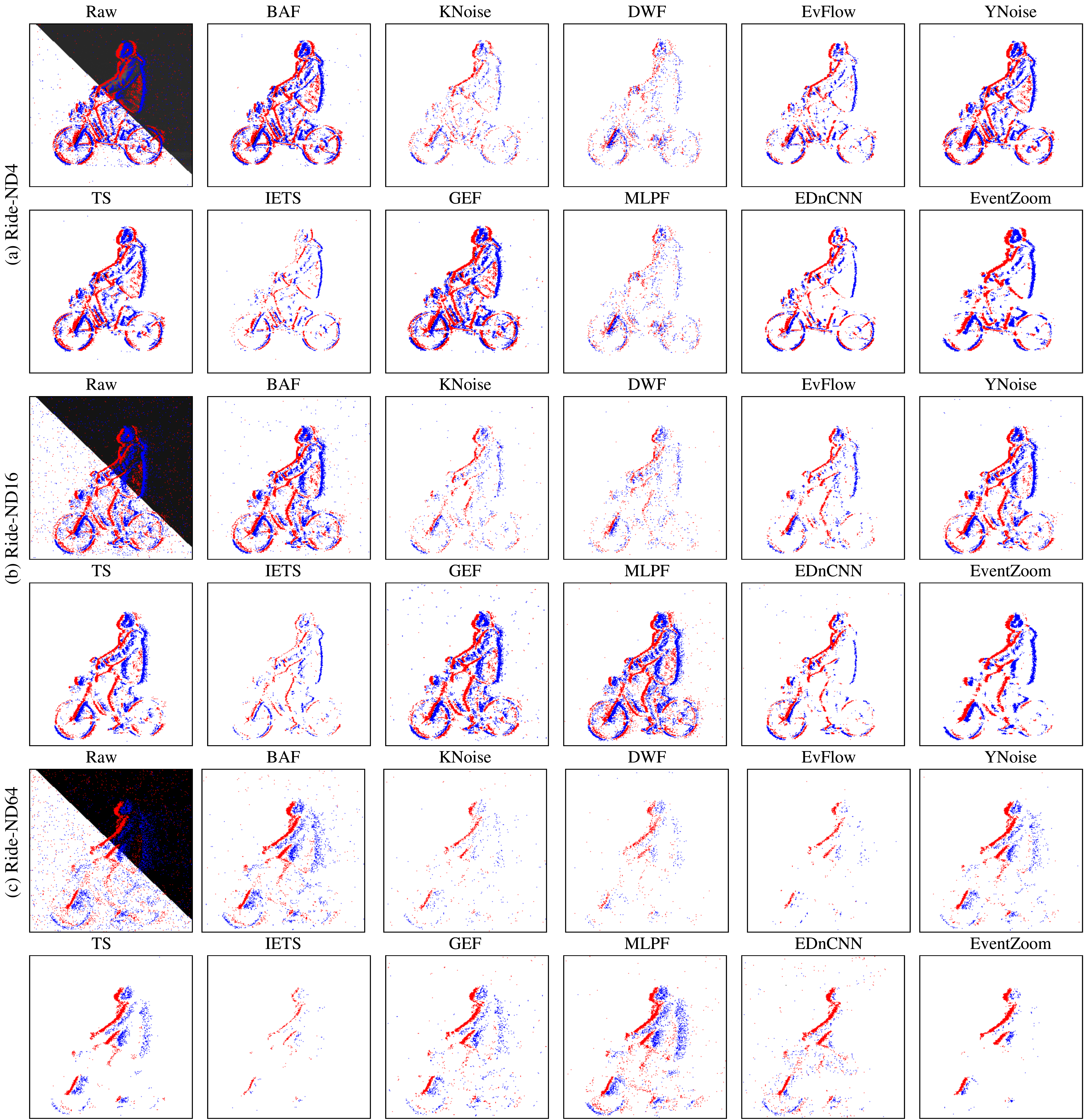}
    \caption{Visual comparison of different denoising algorithms on multiple noise levels of the E-MLB dataset. (a)-(c) contain a cyclist who maintains the same movement at an almost consistent distance from the event camera; as the noise level increases, the edges become more blurred, and details disappear.}
    \label{fig:CrossNoiseLevel}
\end{figure*}

\subsubsection*{\bf Qualitative Evaluation} First, we visualize the denoising results of different algorithms in some challenging ND1 sequences to determine the performance boundary of each denoiser, as shown in \cref{fig:CrossSceneDaytime} (Daytime) and \cref{fig:CrossSceneNight} (Night).

Generally, BAF remains noisy after denoising because it only performs simple density statistics on the event stream but can preserve the edges from being damaged. Although KNoise and DWF follow the same denoising principles as BAF, they perform inferiorly in some complex structural scenes such as \cref{fig:CrossSceneDaytime} (a). This is because they limit the memory space, resulting in a large number of valid events being filtered out rapidly owing to memory limitations and the high noise density. EvFlow performs well when the scene motion type is limited to a single object motion such as \cref{fig:CrossSceneDaytime} (a). To some extent, YNoise and TS perform similarly, but they have distinguished denoising strategies. In detail, TS removes as much noise as possible by local plane fitting, which may damage the texture and details. In contrast, YNoise is a kind of kernel density estimation method that can preserve more structural information. However, YNoise may become invalid in some high-intensity mono-polar noise sequences such as in \cref{fig:CrossSceneDaytime} (a); additionally, YNoise actually costs much more human labor on adjustment. As a denoising method for fast corner detection, IETS destroys the distribution of real events. Although it is highly suppressed in background activities, the edge of the target is no longer obvious. Benefiting from the addition of APS information, the GEF output contains sharp edges and rich texture details, as shown in \cref{fig:CrossSceneDaytime} (a) and (b). However, when the quality of the APS image is poor, the quality of output events also decreases drastically. For example, in \cref{fig:CrossSceneNight} (a), we can see that GEF cannot generate a reasonable event distribution because motion blur occurs.

For neural networks, since MLPF has a simple structure (only 2 hidden layers), it can be difficult to extract global information, resulting in poor performance in complicated scenes such as \cref{fig:CrossSceneDaytime} (a). However, MLPF has the lowest computational and parameter costs compared with other networks. Although EDnCNN can preserve edges well, it loses some texture information of the scene. In addition, we can see EDnCNN's weakness in dealing with object motion in \cref{fig:CrossSceneNight} (a) or extreme noise environment in \cref{fig:CrossSceneNight} (b). Comparatively, EventZoom has more robust performances in various sequences; however, it can cause time and pixel jittering in each event, such as in \cref{fig:CrossSceneDaytime} (a).

Second, we present the denoising results in the same scene with different lighting conditions in \cref{fig:CrossNoiseLevel}. As seen in \cref{fig:CrossNoiseLevel}, the performance of all methods decreases as the noise level increases, and most fail in ND64 sequences. For BAF, NN and KNoise, their denoising sequences are contaminated as the noise level increases, but they have the least computational consumption. As GEF switches to self-guide mode due to the poor quality of the APS frames, it only performs well under moderate noise levels (ND4 and ND16). When the noise level continues to increase (ND64), GEF loses many real events. TS, IETS and YNoise outperform the other methods at medium noise levels and below in \cref{fig:CrossNoiseLevel} (a)-(b), but IETS loses performance at extreme noise conditions in \cref{fig:CrossNoiseLevel} (c). With regard to EDnCNN and EventZoom, each has its own merits: EDNcNN performs well in texture preservation, while EventZoom can retain more edge information. However, both of them may have undesirable performance in some high-noise scenes, specifically compared with TS and YNoise in \cref{fig:CrossNoiseLevel} (c).

\subsubsection*{\bf Comparison between ESR and RPMD} 
\begin{figure}
    \centering
    \includegraphics[width=\linewidth]{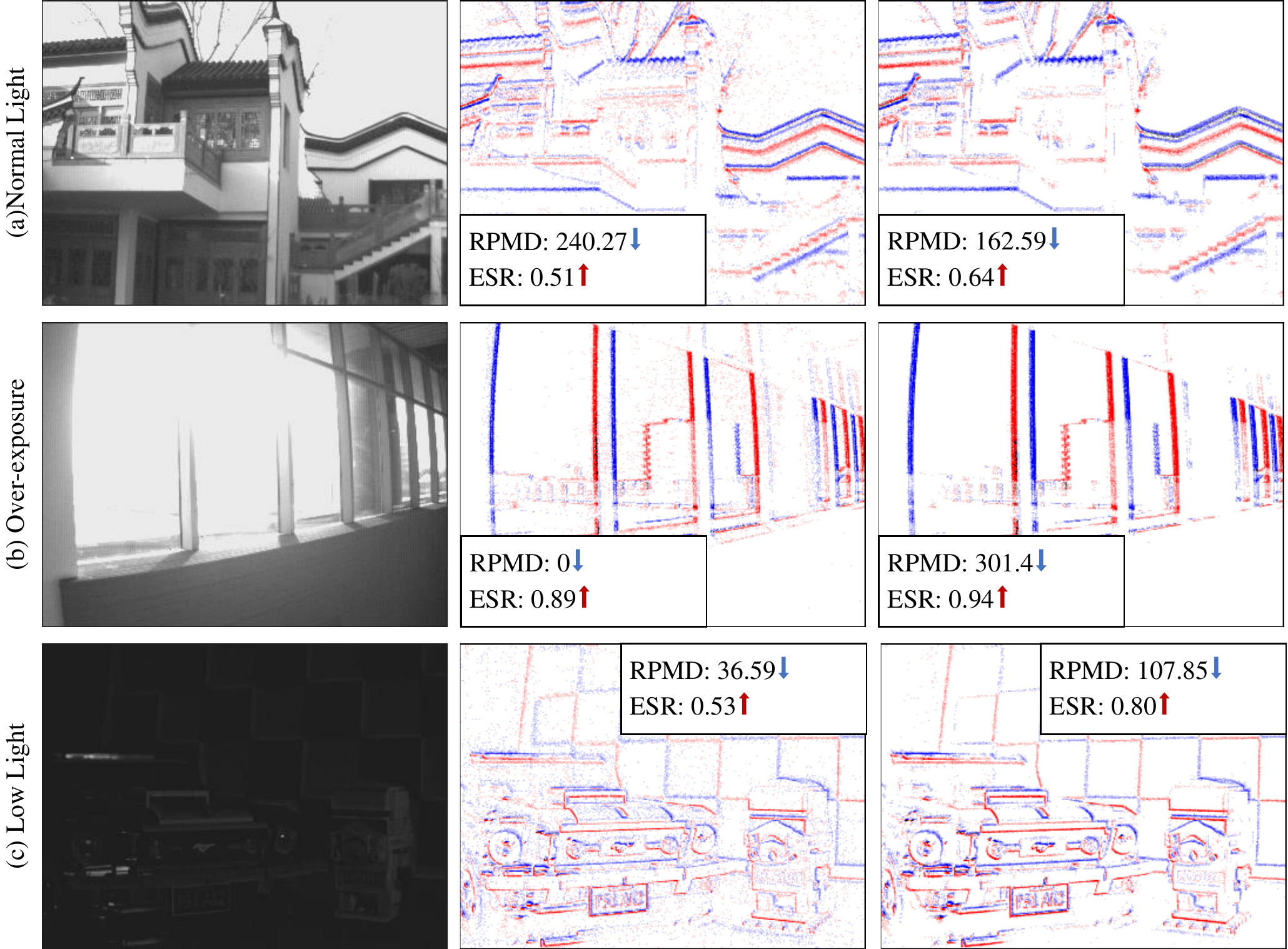}
    \caption{The comparison of ESR with RPMD. (a) shows a normal light sequence, and both methods give reliable scores. (b)-(c) provide an overexposed and a low light sequence correspondingly, which leads to the unexpected results of RPMD, but the proposed ESR still works.}
    \label{fig:RPMD_ESR}
\end{figure}

The proposed ESR in this paper is the first nonreference event denoising metric, which solves the difficulties in obtaining real event labels. In \cref{fig:RPMD_ESR}, we provide a comparison with another common public reference metric, RPMD. Since other methods are not suitable for evaluation on our E-MLB dataset (PSR/PNR and NIR require manual labeling, EDP and ROC require noise-free reference), we do not provide them here.

We visualize the MESR and RPMD scores on 3 representative scenes under normal light, overexposed, and low light conditions in \cref{fig:RPMD_ESR}. As seen, the denoised event frames look better for all the sequences by human perception. However, RPMD fails to give a better score under overexposed and low light sequences. This is because the correct calculation of RPMD requires high-quality and properly aligned APS and IMU data, which is not always met when the event camera is used in the real world. In comparison, the proposed ESR is not dependent on additional information sources and faithfully represents the noise level under all circumstances. Overall, our metric could ignore the restriction to lighting conditions and give more reasonable scores.

\section{Discussion}
\label{section:Discussion}

In this paper, we propose a large-scale event denoising dataset E-MLB and a nonreference event denoising metric ESR for the first time. The scale of E-MLB is 12 times larger than the largest existing event-denoising dataset and rich in noise levels and scene types. The ESR represents the intrinsic property of events without needing any other information sources. With the proposed dataset and event denoising metric, we conduct extensive experiments with 11 state-of-the-art denoising methods and present a comparative analysis on event denoising.

However, there are still some limitations that need to be noted. As discussed in \cite{simulator-v2e}, the dominant event noise source changes from random photocurrent fluctuation to structural junction leakage current as light intensity increases. However, due to the complexity of the scene light sources, we do not discuss and classify the sources of various noise types in our proposed dataset. The proposed metric is easily affected by hot pixels, which are events emitted on some pixels at abnormally high rates. Therefore, we recommend eliminating these unexpected pixels in preprocessing. In future work, we will work on solving the above problems. We hope all these contributions can contribute to the event community to advance future research on event denoising.

\bibliographystyle{IEEEtran}
\bibliography{MLB}

\newpage

\vfill

\end{document}